\documentclass[10pt,twocolumn,letterpaper]{article}

\usepackage{cvpr}              
\definecolor{cvprblue}{rgb}{0.21,0.49,0.74}
\usepackage[pagebackref,breaklinks,colorlinks,allcolors=cvprblue]{hyperref}
\usepackage{multirow}
\usepackage{amsfonts}
\usepackage[table]{xcolor}
\usepackage{booktabs}
\usepackage{pifont}
\usepackage{nicematrix}
\usepackage{amsmath}
\usepackage{algpseudocode}
\usepackage{booktabs}
\usepackage[table]{xcolor}
\usepackage{algorithm}
\usepackage{amsthm}
\usepackage{amsmath}

\usepackage{algpseudocode}
\usepackage{bibentry}

\def\confName{CVPR}
\def\confYear{2026}

\title{What Your Features Reveal: Data-Efficient Black-Box Feature Inversion Attack for Split DNNs}

\author{Zhihan Ren, Lijun He$^\dag$, Jiaxi Liang, Xinzhu Fu, Haixia Bi, Fan Li\\
Xi'an Jiaotong University\\
Xi'an, 710049, China\\
{\tt\small \{renzh,liangjiaxi,xinzhufu\}@stu.xjtu.edu.cn, \{lijunhe,haixia.bi,lifan\}@mail.xjtu.edu.cn}
\and
}

\begin{document}
\maketitle
\definecolor{myhighlight}{gray}{0.9}
\begin{abstract}
Split DNNs enable edge devices by offloading intensive computation to a cloud server, but this paradigm exposes privacy vulnerabilities, as the intermediate features can be exploited to reconstruct the private inputs via Feature Inversion Attack (FIA).
Existing FIA methods often produce limited reconstruction quality, making it difficult to assess the true extent of privacy leakage.
To reveal the privacy risk of the leaked features, we introduce FIA-Flow, a \textbf{black-box} FIA framework that achieves high-fidelity image reconstruction from intermediate features. 
To exploit the semantic information within intermediate features, we design a Latent Feature Space Alignment Module (LFSAM) to bridge the semantic gap between the intermediate feature space and the latent space.
Furthermore, to rectify distributional mismatch, we develop Deterministic Inversion Flow Matching (DIFM), which projects off-manifold features onto the target manifold with \textbf{one-step inference}.
This decoupled design simplifies learning and enables effective training with \textbf{few image–feature pairs}.
To quantify privacy leakage from a human perspective, we also propose two metrics based on a large vision-language model. 
Experiments show that FIA-Flow achieves more faithful and semantically aligned feature inversion across various models (AlexNet, ResNet, Swin Transformer, DINO, and YOLO11) and layers, revealing a more severe privacy threat in Split DNNs than previously recognized.
\end{abstract}     
\section{Introduction}
\label{sec:intro}
\begin{figure}[t]
    \centering
    \includegraphics[width=0.9\linewidth]{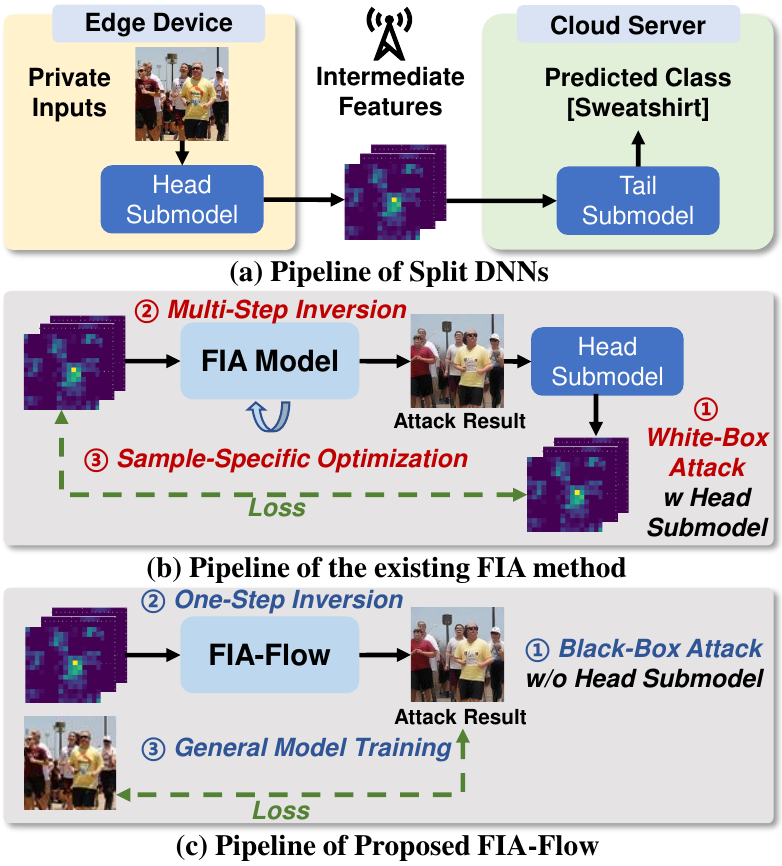}
    \caption{
    (a) The pipeline of Split DNNs, which exposes intermediate features and creates an attack surface.
    (b) Existing FIA methods achieve inversion via white-box, sample-specific iterative feature matching for each input.
     (c) In contrast, FIA-Flow is trained once on a proxy dataset, learning to perform fast one-step inversion for any unseen input.
     }
    \label{fig-intro}
\end{figure}

Deep neural networks (DNNs) have demonstrated remarkable performance in various applications, including autonomous driving \cite{qian2024nuscenes,chen2024end}, smart security \cite{pang2021deep,liu2024vadiffusion,liu2025dependency}, and smart mobile devices \cite{li2018deeprebirth,shiraz2012review,he2024unsupervised,ren2023context}. 
Although performance gains are largely driven by increasing model scale and architectural complexity \cite{kaplan2020scaling}, the resulting computational demands render on-device implementation impractical for resource-constrained edge devices.
To offload the majority of computation to cloud servers, Split DNNs have been proposed \cite{kang2017neurosurgeon,teerapittayanon2017distributed}, which divide a DNN into a lightweight head submodel on edge devices and a computationally intensive tail submodel on cloud servers, as shown in Fig. \ref{fig-intro}(a).
The effectiveness of split computing critically depends on identifying optimal partition points that balance edge computation, cloud processing, and communication overhead across different model architectures and edge device capabilities.

Beyond computational efficiency, split computing is often regarded as a privacy-preserving technique, as raw input data remains on the client's local device \cite{teerapittayanon2016branchynet,kang2017neurosurgeon,karjee2022split,ahuja2023neural,mubark2024asap}. 
However, with the enhanced capabilities of image generation \cite{karras2019style,karras2020analyzing}, this assumption requires serious reconsideration.
While classic model inversion attacks (MIA) \cite{kahla2022label,ye2023c2fmi,zhou2024model,liu2024prediction,li2025head,li2025sample} exploit final model outputs to reconstruct \textbf{training data}, split computing exposes intermediate feature representations during transmission, creating a more direct and vulnerable attack surface.
The potential adversaries include malicious attackers intercepting transmitted features and curious cloud servers analyzing user features beyond their intended computational scope.

This gives rise to the feature inversion attack (FIA), which aims to reconstruct the \textbf{original input images} from intermediate features. 
Despite growing research interest in this threat model, existing FIA methods face three limitations (shown in Fig. \ref{fig-intro}(b) and Table \ref{table-intro}): 
\textbf{(i) White-box assumptions}: Most existing approaches assume white-box access to model architectures and weights \cite{mahendran2015understanding,dmitry2020deep,rojas2022inverting,liang2025analysis,lei2025drag}, limiting their generalization to diverse real-world split computing deployments.
\textbf{(ii) Heavy data dependence}: Learning-based methods \cite{chen2024dia,zhang2024unlocking} typically require extensive training datasets with paired features and ground-truth images, which are difficult to obtain in realistic scenarios. 
\textbf{(iii) High computational cost}: Optimization-based approaches \cite{mahendran2015understanding,dmitry2020deep,zhang2024unlocking,liang2025analysis,lei2025drag} require thousands of iterative optimization steps per sample, making real-time attacks infeasible and easily detectable due to excessive query patterns.

\begin{table}[t]
\caption{Key characteristics and capabilities of various FIA methods. \textsuperscript{\dag} and \textsuperscript{\ddag} denote the different settings of DMB.}
\label{table-intro}
\centering
\setlength{\tabcolsep}{0.1mm}
\definecolor{myhighlight}{gray}{0.9}
\renewcommand\arraystretch{0.8}
\begin{tabular}{lccc}
\toprule
\parbox{1.2cm}{Method} & 
\parbox{1.6cm}{\centering Black-Box Attack} & 
\parbox{1.6cm}{\centering Efficient Inference}& 
\parbox{3.1cm}{\centering Model Applicability \\ (Training Numbers)
}\\ 
\midrule
M\&V \cite{mahendran2015understanding} &\ding{56} & \ding{56}&Sample-Specific\\
DIP \cite{dmitry2020deep} &\ding{56} & \ding{56}&Sample-Specific\\
SG-DIP \cite{liang2025analysis} &\ding{56} & \ding{56}&Sample-Specific\\
DRAG \cite{lei2025drag}& \ding{56}&\ding{56} & Sample-Specific\\
AR \cite{rojas2022inverting}&\ding{56} & \ding{52} & General ($1.28$M)\\
DIA \cite{chen2024dia}& \ding{52}& \ding{56}& General ($40,000$)\\
DMB\textsuperscript{\dag} \cite{zhang2024unlocking}& \ding{52}& \ding{56} & General ($4,096$)\\
DMB\textsuperscript{\ddag} \cite{zhang2024unlocking}& \ding{56}& \ding{56} & Sample-Specific\\
\rowcolor{myhighlight} FIA-Flow &\ding{52} & \ding{52}& General ($< 4,096$)\\
\bottomrule
\end{tabular}
\end{table}

To address these limitations, we propose \textbf{FIA-Flow}, a \textit{black-box} FIA framework built on an alignment-refinement paradigm that simultaneously achieves \textit{one-step inference} and \textit{data-efficient training}, as shown in Fig. \ref{fig-intro}(c).
Specifically, the alignment stage employs a Latent Feature Space Alignment Module (LFSAM) that bridges the semantic gap between task-specific intermediate features and generative latent spaces.
LFSAM progressively fuses multi-channel information and adapts to diverse network layers and architectures, mapping the intermediate feature into a structurally aligned latent representation.
Furthermore, the refinement stage develops the Deterministic Inversion Flow Matching (DIFM), inspired by flow-matching (FM) \cite{lipman2022flow}. 
Unlike conventional generative models \cite{liu2022flowstraightfastlearning,zhu2024flowie}, DIFM learns a deterministic vector field to project these coarsely aligned features onto the natural data manifold, correcting distributional mismatch and recovering fine-grained visual details.
Crucially, FIA-Flow operates in a black-box setting, requiring only query access to intermediate features, and can be trained effectively with a small collection (fewer than $4,096$ image-feature pairs, i.e., $<0.32\%$ of ImageNet-1K), making it highly practical for real-world split computing scenarios.
The main contributions are as follows:
\begin{itemize}
    \item \textbf{Black-box FIA framework}: We propose FIA-Flow, a black-box FIA framework that can eliminate iterative optimization without requiring access to the victim model, thereby revealing the risks present in split computing.
    \item \textbf{Data-efficient alignment-refinement strategy}: We decouple the FIA task into a two-stage paradigm combining LFSAM for cross-space feature mapping and DIFM for distributional correction with few samples.
    \item \textbf{One-Step Inference via DIFM}: We develop DIFM that learns a deterministic vector field to enable high-fidelity reconstruction in a single forward pass, eliminating the iterative optimization of optimization-based methods and multi-step sampling of diffusion-based FIA models. 
\end{itemize}

\section{Method}

\begin{figure*}[t]
    \centering
    \includegraphics[width=0.9\linewidth]{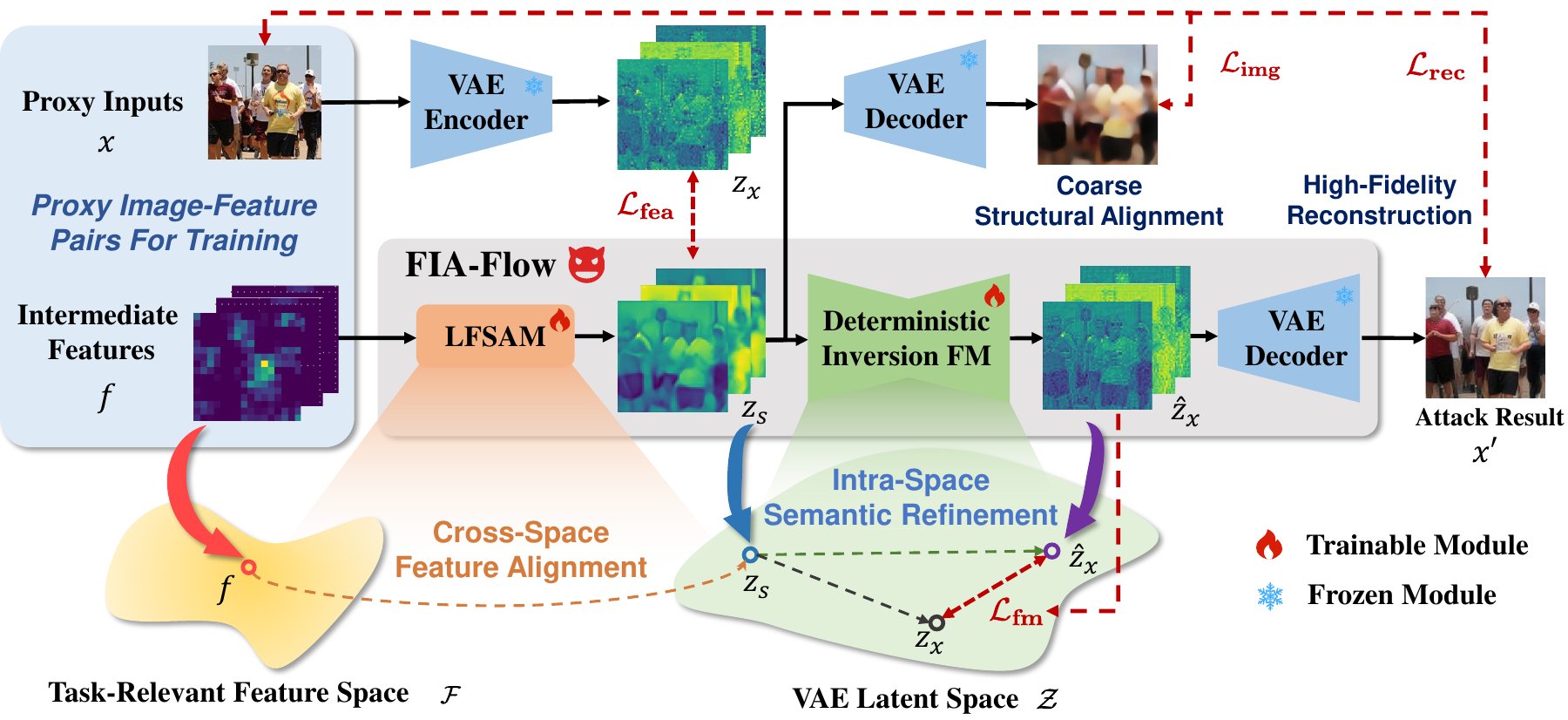}
    \caption{The pipeline of FIA-Flow. The method reconstructs a private image $x$ from the corresponding intermediate features $f$. It first maps $f$ to a latent code $z_s$ by the Latent Feature Space Alignment Module, then uses the Deterministic Inversion Flow Matching module to refine it into $\hat{z}_x$. Finally, the attack image $x'$ is obtained by a pre-trained VAE decoder from $\hat{z}_x$.}
    \label{fig-pipe}
\end{figure*}
\subsection{Overview and Problem Formulation}
Let $M: \mathcal{X} \to \mathcal{F}$ denote the head submodel of the victim Split DNN system, where $\mathcal{X} \subseteq \mathbb{R}^{H \times W \times C}$ is the space of private input images. 
For a given input $x \in \mathcal{X}$, the model $M$ produces an intermediate feature $f=M(x)$ at a specific split layer, where $f \in \mathcal{F} \subseteq \mathbb{R}^{D_f}$ and $D_f$ is the feature dimension.
The primary objective of FIA is to learn an inversion mapping $G: \mathcal{F} \to \mathcal{X}$ that can reconstruct the original input $x'$ from its corresponding feature $f$, such that the reconstructed image $x' = G(f) \approx x$ is perceptually and semantically indistinguishable from the original input $x$. 

Our attack operates under a black-box assumption, where the architecture and parameters of $M$ are unknown. 
We can only query to obtain a set of image-feature pairs $\mathcal{D} = \{ (x_i, f_i) \}_{i=1}^{N}$ for training.
To achieve this, FIA-Flow adopts an alignment-refinement strategy, as shown in Fig. \ref{fig-pipe}.
We decouple the complex inversion mapping $G$ into a two-stage process: a structural alignment stage and a semantic refinement stage, which can be formulated as:
\begin{align}
x'=G(f) = \text{Dec}(G_{\text{refine}}(G_{\text{align}}(f)) )
\end{align}
$\text{Dec}:\mathcal{Z} \to \mathcal{X}$ denotes the decoder of Variational Autoencoder (VAE)  \cite{kingma2013auto}. 
The alignment stage $G_{\text{align}}: \mathcal{F} \to \mathcal{Z}$ establishes structural correspondence by aligning the task-relevant feature spaces and latent space of the VAE. 
However, this alignment primarily yields an off-manifold representation, which lacks the semantic richness.
Therefore, the refinement stage $G_{\text{refine}}: \mathcal{Z} \to \mathcal{Z}$ performs intra-space semantic enhancement, correcting the distributional mismatch to ensure high-fidelity inversion.

\subsection{Latent Feature Space Alignment Module}
\paragraph{Motivation and Objective} 
A fundamental challenge in FIA arises from the space gap between $\mathcal{F}$ and the latent space $\mathcal{Z}$. 
Since $\mathcal{F}$ is task-specific and optimized for classification rather than synthesis, its structure is inherently incompatible with the manifold of $\mathcal{Z}$ \cite{yosinski2014transferable}.
Therefore, a direct mapping from feature $f$ to image $x$ is ill-posed.
To bridge this gap, we propose the LFSAM to transform a given intermediate feature $f$ into a latent tensor $z_s=G_{\text{align}}(f)$, which is designed to be both dimensionally compatible and structurally aligned with the latent space of VAE. 
The VAE is selected for its continuity and structured latent space, offering an ideal manifold for stable and coherent refinement \cite{doersch2016tutorial}. 
Meanwhile, the low-dimensional latent space reduces the complexity of the hypothesis class, making alignment learning easier to generalize under few-sample conditions \cite{yang2024few}. 
This enables FIA-Flow to effectively learn and extrapolate robustly to unseen features.

\paragraph{Cross-Space Feature Alignment}  
LFSAM comprises a learned upsampling module, a backbone, and a Feature Aggregation Network (FAN) to synthesize a comprehensive latent representation. 
To accommodate features with varying resolutions across different network layers, we employ a PixelShuffle-based spatialization layer $PS: \mathbb{R}^{(r^2  C_{in}\times  H_{in} \times W_{in})} \to \mathbb{R}^{(C_{in} \times r  H_{in} \times r W_{in})}$. 
Unlike standard interpolation, this operation provides a learned transformation that unfolds channel-encoded spatial information into an explicit geometric grid. 

The backbone $B(\cdot)$ processes $f$ through a hierarchical encoder, which extracts a set of feature maps $\{e_1,e_2, \ldots, e_L\}$. 
Its corresponding decoder reconstructs the output progressively from the deepest feature level. 
Crucially, at each stage, the decoder integrates features from the corresponding encoder stage via skip connections, a process formulated as $d_{i+1}=\mathcal{D}(concat(d_i,e_{i}))$. 
To capture global context and long-range spatial dependencies, we embed self-attention mechanisms within the backbone layers, producing $F_{d}=B(f)$. 
Meanwhile, FAN projects each $e_i$ into a shared space via $1\times 1$ convolutions $\phi_i$ then concatenates and fuses them: $F_{fan}=\text{Conv}_{\text{fuse}}(\text{concat}_{i=1}^L(\phi_i(e_i)))$.
The final aligned feature is:
\begin{equation}
z_s = \text{Conv}_{\text{out}}(F_{d} + F_{fan}),
\end{equation}
which serves as a structural alignment feature for the subsequent refinement stage.

\subsection{Deterministic Inversion Flow Matching}
\label{sec:fm} 
\paragraph{Motivation and Objective}
With the space-aligned latent feature $z_s$ obtained from LFSAM, we aim to generate a photorealistic inversion image $x'$ that closely resembles the private input $x$. 
A naive approach involves directly decoding the aligned feature $z_s$ using a pre-trained VAE decoder $x' = \text{Dec}(z_s)$. 
However, experimental results demonstrate that this straightforward method produces suboptimal results with severe blurriness and semantic inconsistencies.

The core issue is a distributional mismatch between the aligned features and the natural data manifold. 
Although LFSAM ensures that $z_s$ conforms to the dimensional requirements of the VAE latent space, it fails to guarantee that $z_s$ follows the same distribution as the authentic latent feature generated by the VAE encoder from natural images.
Since $z_s$ originates from a task-specific feature transformation, it likely resides in off-manifold regions of the latent space $\mathcal{Z}$. 
The VAE decoder is trained exclusively on on-manifold samples, cannot interpret out-of-distribution inputs, resulting in degraded reconstruction quality.
Therefore, we propose the DIFM to enhance semantic expressiveness based on the previous structural alignment.
\paragraph{Intra-Space Feature Enhancement}
To overcome the limitations of direct decoding, we reframe $z_s$ as a high-quality starting point for a generative process rather than a final latent feature. 
We employ the DIFM to learn a deterministic vector field $v_\theta(z,t)$ that transforms the distribution of our structurally-aligned features $p_0 = p(z_s)$ to the target data distribution $p_1 = p(z_x)$, where $z_x=\text{Enc}(x)$.
This approach adapts the standard FM framework by replacing the conventional Gaussian prior $p_0' = \mathcal{N}(0, I)$ with our meaningful initializations $p(z_s)$.

Specifically, we define a linear interpolation path between the starting point $z_s$ and its corresponding target $z_x$ as $z_t = t \cdot z_x +(1-t) \cdot z_s $ for $t\in [0,1]$. 
DIFM is trained to approximate this target field $u_t={dz_t}/{dt}=z_x-z_s$. 
Once trained, this learned vector field defines the trajectory of each point via the probability flow ordinary differential equation (ODE), $d\hat{z}_t/{dt}=v_{\theta}(\hat{z}_t,t)$, and the continuity equation describes its distributional evolution:
\begin{equation}
\partial_t p_t(z) + \nabla_z \cdot (p_t(z) v_\theta(z, t)) = 0.
\end{equation}
This equation formalizes the desired behavior of $v_\theta(z,t)$, ensuring it guides the population of points from the initial distribution $p_0$ to the target data distribution $p_1$.
Since LFSAM already produces $z_s$ close to $z_x$, the learned vector field is simple, allowing us to replace an expensive ODE solver with a single forward Euler step from $t=0$ to $t=1$:
\begin{equation}
\hat{z}_x =\hat{z}_1= z_s +v_\theta(z_s, t=0).
\end{equation}
The final inversion image $x'$ is decoded by the VAE: $x'=\text{Dec}(\hat{z}_x)$. 
By conditioning the generative process on a meaningful initialization, our strategy effectively transforms a complex generation task into a residual correction problem. 
This simplifies the learning dynamics of the vector field, reducing the data requirements, thereby enabling high-fidelity inversion even with limited training samples.

\subsection{Training Strategy}
\label{sec:training}
We adopt a two-stage training paradigm for the FIA task. This decoupled approach is designed first to establish a space alignment and then to optimize the generative model. \\
\textbf{Stage 1: Training the LFSAM.}
In the first stage, we train the LFSAM to learn a mapping from the task-relevant input features $f$ to the VAE latent space. Our objective is to ensure that the LFSAM produces structured features $z_s$ that closely approximate the ground truth (GT) VAE latent feature $z_x$ of the corresponding images $x$.
We employ a pre-trained, frozen VAE encoder to obtain target latent codes $z_x = \text{Enc}(x)$. 
To ensure feature space alignment, we minimize the L2 distance between the $z_s$ and $z_x$:
\begin{equation}
\label{eq-fea}
\mathcal{L}_{\text{fea}} = \mathbb{E}_{(x, f) \sim \mathcal{D}} \left[ \| z_s - z_x \|_2^2 \right].
\end{equation}

To enforce perceptual coherence, we apply an image-domain reconstruction loss. We decode the generated latent $z_s$, and minimize the L1 distance to the GT image $x$:
\begin{equation}
\label{eq-img}
\mathcal{L}_{\text{img}} = \mathbb{E}_{(x, f) \sim \mathcal{D}} \left[ \| \text{Dec}(z_s) - x \|_1 \right].
\end{equation}

The total loss for Stage 1 is the sum of these two losses:
\begin{equation}
\mathcal{L}_{\text{s1}} = \mathcal{L}_{\text{fea}} + \mathcal{L}_{\text{img}}.
\end{equation}
\begin{figure*}[t]
    \centering
    \includegraphics[width=0.95\linewidth]{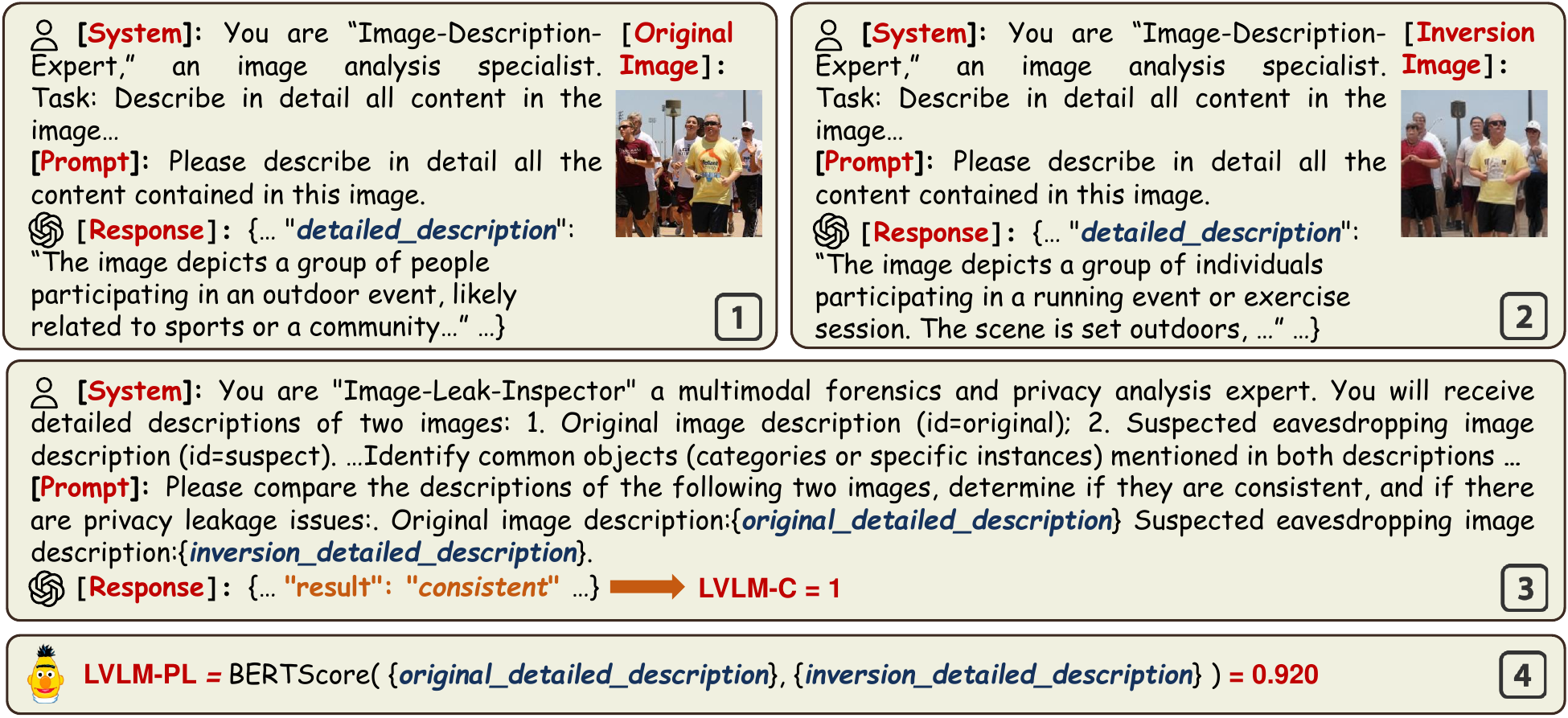}
    \caption{An illustration of LVLM-C and LVLM-PL evaluation. \ding{172} The LVLM is prompted to describe the original image. \ding{173} The LVLM is then prompted to describe the inversion image. \ding{174} The LVLM compares these two descriptions to ascertain if the same object is identified. A consistent result yields the LVLM-C value of 1. \ding{175} LVLM-PL is obtained by computing the BERTScore \cite{zhang2019bertscore} between the two descriptions. 
    }
    \label{fig-lvlm}
\end{figure*}

Stage 1 ensures that the LFSAM learns a meaningful projection into the VAE latent space, providing a solid foundation for the subsequent stage. \\
\textbf{Stage 2: Training the DIFM.}
In the second stage, we freeze the LFSAM and train the DIFM, which takes the precomputed features $z_s$ as a starting point and learns to generate the final image $x'$. 
The training objective for this stage is a combination of two losses:

\begin{enumerate}
    \item \textbf{Flow Matching Loss ($\mathcal{L}_{\text{fm}}$):} 
    This is a regression loss that minimizes the L2 distance between the model's predicted vector field $v_\theta(z_t,t)$ and the target vector field $u_t$:
    \begin{equation}
    \mathcal{L}_{\text{fm}} = \mathbb{E}_{t \sim \mathcal{U}[0,1], (x,f) \sim \mathcal{D}} \left[ \| v_\theta(z_t,t) - u_t \|_2^2 \right].
    \end{equation}

    \item \textbf{Reconstruction Loss ($\mathcal{L}_{rec}$):} To ensure that the final output $x'$ is perceptually and semantically faithful to the original image $x$, we apply a reconstruction loss directly in the image space. 
    This loss is a combination of Learned Perceptual Image Patch Similarity (LPIPS) \cite{zhang2018unreasonable} loss and a pixel-wise L1 loss:
    \begin{equation}
    \label{eq-rec}
    \mathcal{L}_{\text{rec}} = \mathbb{E}_{(x,f) \sim \mathcal{D}} \left[\ \mathcal{L}_{\text{LPIPS}}(x', x) + \mathcal{L}_{\text{L1}}(x', x) \right].
    \end{equation}
\end{enumerate}

The final loss for Stage 2 is the sum of these two losses:
\begin{equation}
\label{eq-all}
\mathcal{L}_{\text{s2}} = \mathcal{L}_{\text{fm}} + \mathcal{L}_{\text{rec}}.
\end{equation}

\section{Experiments}
\subsection{Datasets and Metrics}
Our experiments were conducted on a subset of ImageNet-1K \cite{deng2009imagenet}. Specifically, we randomly sample only 4,096 images ($<0.32 \%$) from the training set for training and 1,000 images from the validation set for testing.

We employ a comprehensive set of Image Quality Assessment (IQA) metrics. For full-reference IQA, we use the Peak Signal-to-Noise Ratio (PSNR), Structural Similarity Index Measure (SSIM), and LPIPS \cite{zhang2018unreasonable}. 
For no-reference IQA, we utilize the Natural Image Quality Evaluator (NIQE) \cite{mittal2012making} and MANIQA \cite{yang2022maniqa}. 
Furthermore, to measure the eavesdropping information accuracy, we assess the inversion image top-1 classification accuracy (\textbf{Acc}) with the GT label of the original image, using ResNet-50 \cite{he2016deep}.
To assess private information leakage, we propose two novel metrics evaluated by Large Vision-Language Models (LVLMs): LVLM-Consistency (\textbf{LVLM-C}) and LVLM-Privacy-Leakage (\textbf{LVLM-PL}).
As shown in Fig. \ref{fig-lvlm}, the LVLM acts as an \textit{Image Description Expert}, generating textual descriptions for both the original and inversion images.
These descriptions are compared by an \textit{Image Leak Inspector} to determine whether they depict the same primary object (LVLM-C) and to compute their semantic similarity via BERTScore \cite{zhang2019bertscore} (LVLM-PL).
Higher LVLM-C and LVLM-PL values indicate that the attacker can extract more detailed private information from the inversion image.
In our implementation, we utilize \textit{gpt-4o-mini} as the LVLM. 
See supplementary materials for the detailed calculation process and an ablation study with other LVLM.
\begin{table*}[t]
\setlength{\tabcolsep}{1.4mm}
\caption{The performance comparison among different FIA methods. Bold indicates the best result of all methods.}
\label{table-compare}
\renewcommand\arraystretch{0.7}
\centering
\begin{tabular}{ccccccccccc}
\toprule
Model & Layer & Method & PSNR $\uparrow$& SSIM $\uparrow$& LPIPS $\downarrow$ & Acc $\uparrow$& LVLM-C $\uparrow$& LVLM-PL $\uparrow$& NIQE $\downarrow$ & MANIQA $\uparrow$\\
  \midrule
 \multirow{6}{*}{AlexNet} & \multirow{6}{*}{F-10} & M\&V & 13.55 & 0.500 & 0.730 & 0.0 & 1.2&0.860 & 5.853 & 0.4303 \\
 & & DIP &15.45 &0.422 &0.585  &16.1 &10.6&0.880 & 5.988 &0.2763 \\
 & & AR & 18.65 & 0.508 & 0.574 & 4.1 & 4.8 & 0.880 & 5.874 &0.3258 \\
  & & SG-DIP & 11.07&0.257 &0.778  & 1.2&3.6 & 0.865 &\textbf{5.603} &0.2950 \\
  & & FIA-Align &20.46 & \textbf{0.607}  &0.620 & 5.7&9.3& 0.883&10.927 &0.2959  \\
 & & \cellcolor{myhighlight}FIA-Flow & \cellcolor{myhighlight}\textbf{20.64}& \cellcolor{myhighlight}0.603&\cellcolor{myhighlight}\textbf{0.405}  &\cellcolor{myhighlight}\textbf{28.8}& \cellcolor{myhighlight}\textbf{16.6}& \cellcolor{myhighlight}\textbf{0.900}& \cellcolor{myhighlight}6.243&\cellcolor{myhighlight}\textbf{0.4956}  \\
 \midrule
\multirow{10}{*}{ResNet-50} & \multirow{5}{*}{L1-2} & M\&V & 13.83 & 0.603 & 0.593 & 13.4 &17.5  &0.903  & 5.392 & 0.4938\\
 & & DIP & 25.73 & 0.706 & 0.236 & 61.0 &  39.9&0.905 & 5.504 & 0.4565 \\
  & & SG-DIP & 27.90&0.754 &0.193  & 65.2& 65.3 &0.922 &5.301 &0.4928 \\
  & & FIA-Align & 29.86 & 0.810&0.157  & 64.3& 70.0&0.923 & 5.136 & 0.5622 \\
 & & \cellcolor{myhighlight}FIA-Flow &\cellcolor{myhighlight}\textbf{30.01} &\cellcolor{myhighlight}\textbf{0.814} &\cellcolor{myhighlight}\textbf{0.100}  &\cellcolor{myhighlight}\textbf{71.3}&\cellcolor{myhighlight}\textbf{70.1} & \cellcolor{myhighlight}\textbf{0.929}& \cellcolor{myhighlight}\textbf{4.408}&\cellcolor{myhighlight}\textbf{0.6131}  \\
\cmidrule{2-11}
 & \multirow{5}{*}{L4-2} & M\&V & 13.55 & 0.504 & 0.851 & 0.0&3.0 &  0.860 & 7.577 & 0.4359\\
 & & DIP & 13.60 & 0.453 & 0.711 &27.3 & 9.4 &0.881 &7.152 & 0.2592 \\
  & & SG-DIP &11.59 &0.309 &0.777  & 8.1&5.0 &0.872&5.603 &0.3189 \\
   & & FIA-Align & \textbf{20.36} & \textbf{0.603} & 0.643  &4.4&6.3&0.878 & 11.309 &0.2969\\
 & & \cellcolor{myhighlight}FIA-Flow & \cellcolor{myhighlight}20.31 & \cellcolor{myhighlight}0.584 & \cellcolor{myhighlight}\textbf{0.397} & \cellcolor{myhighlight}\textbf{36.8} & \cellcolor{myhighlight}\textbf{18.0} & \cellcolor{myhighlight}\textbf{0.902} & \cellcolor{myhighlight}\textbf{5.098} & \cellcolor{myhighlight}\textbf{0.5628} \\
 \midrule
 \multirow{5}{*}{Swin-B}& \multirow{5}{*}{F3-2} & M\&V &14.34 &0.628 &0.541  & 38.1& 38.4& 0.913&6.105 & 0.4465\\
 & & DIP & 21.03&0.735 &0.313  & 61.7& 54.5&0.920 &5.486 &0.4492 \\
  & & SG-DIP&25.15 &\textbf{0.872} &0.191  & 68.5&62.3 & 0.913&5.520 &0.5362 \\
  & & FIA-Align & 26.64& 0.771&0.260  & 53.6& 51.5&0.919 &6.236 &0.4725 \\
   & & \cellcolor{myhighlight}FIA-Flow &
    \cellcolor{myhighlight}\textbf{27.29} &
    \cellcolor{myhighlight}0.780 &
    \cellcolor{myhighlight}\textbf{0.159} &
    \cellcolor{myhighlight}\textbf{70.6} &
    \cellcolor{myhighlight}\textbf{63.2} &
    \cellcolor{myhighlight}\textbf{0.925} &
    \cellcolor{myhighlight}\textbf{4.840} &
    \cellcolor{myhighlight}\textbf{0.5777} \\
\midrule
  \multirow{5}{*}{YOLO11n}& \multirow{5}{*}{M-8} & M\&V &7.59 &0.239 &0.890 &0.3 & 1.2& 0.863&6.715 & 0.4702\\
 & & DIP & 14.09&0.521 &0.572 & 14.6& 18.2&0.897 & 6.796&0.4168 \\
  & & SG-DIP& 14.04&0.518 &0.582 & 12.1& 18.1& \textbf{0.899}&6.696 &0.4124 \\
  & & FIA-Align &20.56 &\textbf{0.612} &0.627 & 4.1& 7.1&0.880 &11.056 & 0.2958\\
   & & \cellcolor{myhighlight}FIA-Flow &
    \cellcolor{myhighlight}\textbf{20.90} &
    \cellcolor{myhighlight}0.608&
    \cellcolor{myhighlight}\textbf{0.437} &
    \cellcolor{myhighlight}\textbf{23.6} &
    \cellcolor{myhighlight}\textbf{23.9} &
    \cellcolor{myhighlight}\textbf{0.899} &
    \cellcolor{myhighlight}\textbf{6.528} &
    \cellcolor{myhighlight}\textbf{0.4968} \\
    \midrule
     \multirow{5}{*}{DINOv2-B}& \multirow{5}{*}{B-11} & M\&V & 13.53&0.477 &0.868 &0.1 &0.7 &0.855 &13.533 &0.3324 \\
 & & DIP & 13.45& 0.493&0.833 &1.3 & 4.9& 0.868& 8.497 &0.3316\\
  & & SG-DIP&12.42 &0.345&0.741 & 17.7 &28.3 & 0.905&5.838 &0.3662 \\
  & & FIA-Align &19.92 &0.619 &0.609 &9.2 & 16.7& 0.890&10.340 &0.2709 \\
   & & \cellcolor{myhighlight}FIA-Flow &
    \cellcolor{myhighlight}\textbf{20.13} &
    \cellcolor{myhighlight} \textbf{0.621}&
    \cellcolor{myhighlight}\textbf{0.411} &
    \cellcolor{myhighlight}\textbf{42.8} &
    \cellcolor{myhighlight}\textbf{30.4} &
    \cellcolor{myhighlight}\textbf{0.909} &
    \cellcolor{myhighlight}\textbf{6.304} &
    \cellcolor{myhighlight}\textbf{0.5079} \\

\bottomrule
\end{tabular}
\end{table*}

\subsection{Implementation Details}
We selected \textit{features.10} (F-10) of AlexNet \cite{krizhevsky2017imagenet}, \textit{layer1.2} (L1-2) and \textit{layer4.2} (L4-2) of ResNet-50 \cite{he2016deep}, \textit{features.3.0.mlp.2} (F3-2) of Swin Transformer (Swin-B) \cite{liu2021swin}, \textit{model.8} (M-8) of YOLO11n \cite{yolo11_ultralytics}, and \textit{blocks.11} (B-11) of DINOv2-B \cite{oquab2023dinov2} as the victim layers and models for FIA.
The DIFM is initialized with the pre-trained weights of Stable Diffusion 2.1 \cite{rombach2022high}. 
To adapt it for the FIA task, we freeze the U-Net in the DIFM and integrate a Low-Rank Adaptation (LoRA) \cite{lora} model with a rank of $r=4$.
For both stages, we set the batch size to $8$ and the learning rate to $0.0001$, with each stage trained for $64,000$ iterations.
All experiments were conducted on NVIDIA A100 GPUs.

\begin{figure*}[t]
    \centering
    \includegraphics[width=0.9\linewidth]{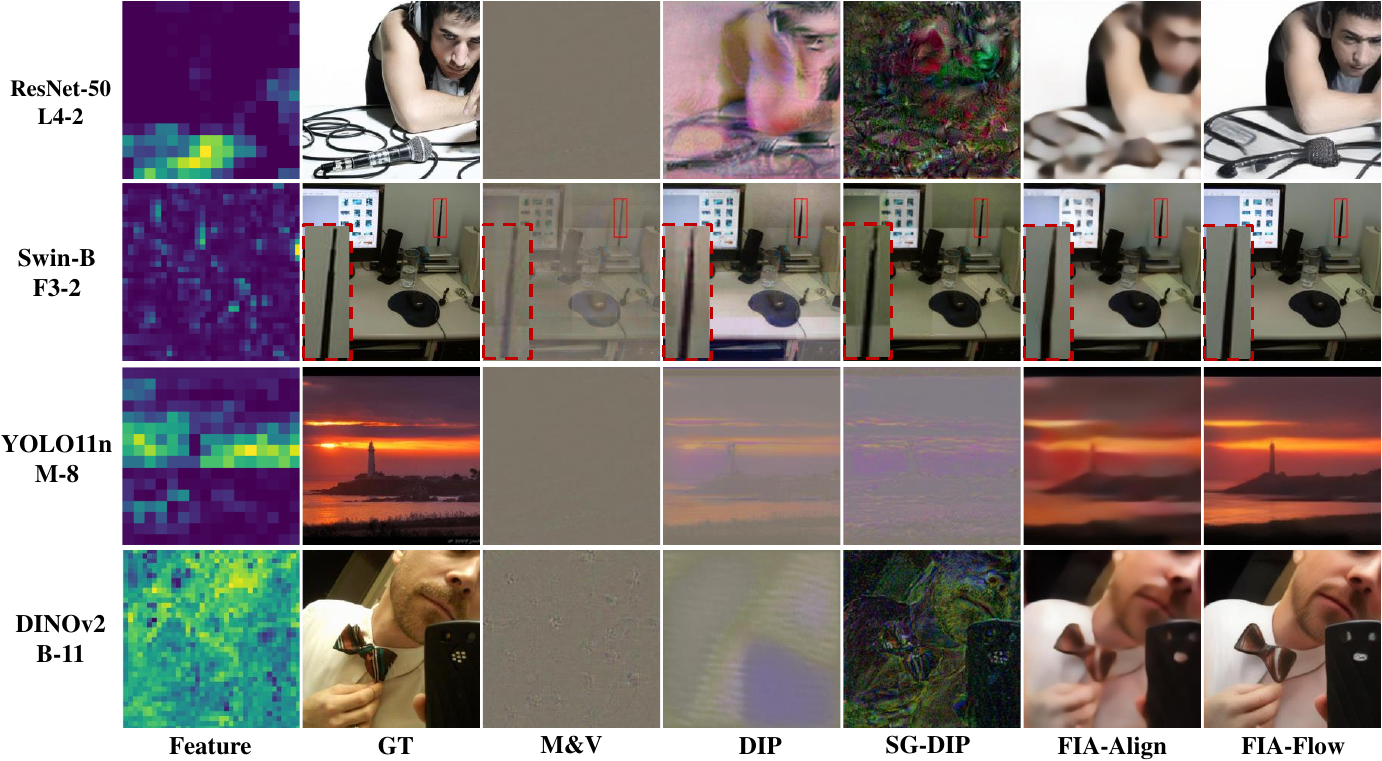}
    \caption{Visualization comparison of different FIA methods on various models. 
    }
    \label{fig-compare}
\end{figure*}
\begin{figure}[t]
    \centering
    \includegraphics[width=0.9\linewidth]{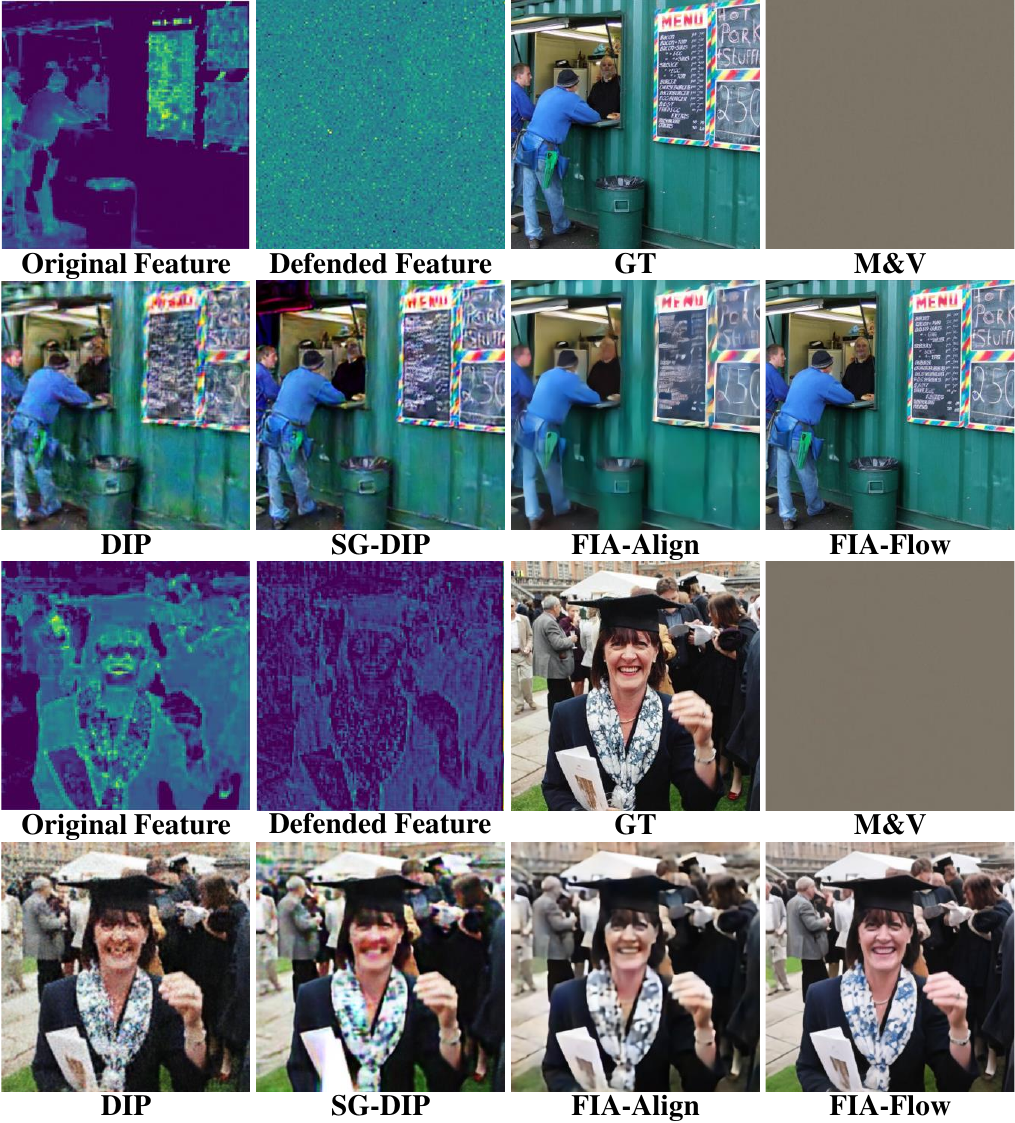}
    \caption{Visualization comparison on different defense mechanisms. Top row: visualizations under the Noise+NoPeek defense \cite{titcombe2021practical}. Bottom row: visualizations under the DISCO defense \cite{singh2021disco}.}
    \label{fig-defend}
\end{figure}

\subsection{Main results}
We compare the proposed FIA-Flow with state-of-the-art FIA methods, including M\&V \cite{mahendran2015understanding}, Deep Image Prior (DIP) \cite{dmitry2020deep}, Adversarially Robust (AR) \cite{rojas2022inverting}, Self-Guided DIP (SG-DIP) \cite{liang2025analysis}. 
Additionally, we compared against a baseline FIA-Align that solely employs LFSAM for feature space alignment, followed by VAE decoding.

\paragraph{Quantitative Results}
Table \ref{table-compare} shows the results of different FIA methods for various victim models and layers.
For AlexNet, FIA-Flow achieves an Acc of 28.8\%, showing a significant advantage over other methods.
For ResNet-50, when dealing with information-rich shallow features (L1-2), FIA-Flow can achieve an outstanding Acc of \textbf{71.3\%}.
This performance remains robust even when dealing with deep features from the L4-2 layer, which typically suffer from substantial information loss. 
While other methods experience a dramatic degradation in image quality, leading to significant drops in both Acc and LVLM-based evaluations, FIA-Flow maintains an Acc of 36.8\% and an LVLM-PL of 0.902. 
Furthermore, experiments on the Swin Transformer model (Swin-B), object detection model (YOLO11n), and foundation model (DINOv2-B) confirm the superiority of FIA-Flow, highlighting its broad applicability and effectiveness across diverse model architectures.

Benefiting from the alignment-refinement strategy, FIA-Flow not only achieves a higher inversion quality on IQA metrics but also exhibits substantially better semantic preservation, as validated by Acc and LVLM-based metrics. 
This proves that FIA-Flow constitutes a more effective and practical privacy threat.

\paragraph{Qualitative Results}
As shown in Fig. \ref{fig-compare}, our FIA-Flow outperforms other methods on both ResNet-50, Swin-B, YOLO11n, and DINOv2. 
While other methods fail or produce blurry results, FIA-Flow can invert images with exceptional clarity, accurately capturing fine details like the face, wireless router, and lighthouse. 
This visually confirms its state-of-the-art performance and robustness across diverse architectures. 
More visual results are available in the Supplementary Materials.

\paragraph{Robustness Evaluation Under Defenses}
To verify the robustness of FIA-Flow under different defense mechanisms, we evaluate all methods against two representative defenses: Noise+NoPeek \cite{titcombe2021practical} and DISCO \cite{singh2021disco}, as shown in Table \ref{ab-defend} and Fig. \ref{fig-defend}.
Under the Noise+NoPeek defense, where Laplacian noise is injected into intermediate features and a NoPeek strategy \cite{vepakomma2020nopeek} is employed to restrict information leakage, FIA-Flow still outperforms other methods. 
Similarly, under the DISCO defense, which suppresses intermediate features, FIA-Flow remains effective, recovering the original image with minimal samples.
This demonstrates that FIA-Flow can effectively bypass defense mechanisms and extract sensitive information, even in a black-box setting, without access to the defense's implementation details and model parameters.

\begin{table}[t]
\caption{Robustness comparison under different defense mechanisms of Split DNNs on the L1–2 layer of ResNet-50.}
\setlength{\tabcolsep}{0.6mm}
\centering
\label{ab-defend}
\renewcommand\arraystretch{0.8}
\begin{tabular}{cccccc}
\toprule
Defense& Methods& PSNR$\uparrow$& Acc$\uparrow$& LVLM-C$\uparrow$& LVLM-PL$\uparrow$\\
\midrule
 \multirow{5}{*}{\shortstack{Noise\\+\\ NoPeek\\ \cite{titcombe2021practical}}}& M\&V&13.56&0.0 & 2.1&0.861\\
  &  DIP&21.87 &26.9& 41.5&0.921 \\
   & SG-DIP&21.69&53.3&49.1&0.919\\
 & FIA-Align&26.05 &38.3 & 45.8&0.911\\
 & \cellcolor{myhighlight}FIA-Flow & \cellcolor{myhighlight}\textbf{27.70} & \cellcolor{myhighlight}\textbf{62.2} & \cellcolor{myhighlight}\textbf{55.0} & \cellcolor{myhighlight}\textbf{0.922} \\
   \midrule
  \multirow{5}{*}{\shortstack{DISCO\\ \cite{singh2021disco}}}&  M\&V
&13.57&0.1&1.0&0.860\\
  & DIP&
\textbf{27.10}&35.9&39.6&0.914\\
  & SG-DIP
& 26.02&43.7&39.8 &0.910\\
 & FIA-Align&
26.49 &37.4 &38.7 &0.913\\
 & \cellcolor{myhighlight}FIA-Flow & \cellcolor{myhighlight}26.75 & \cellcolor{myhighlight}\textbf{59.0} & \cellcolor{myhighlight}\textbf{44.8} & \cellcolor{myhighlight}\textbf{0.916} \\
 \bottomrule
\end{tabular}
\end{table}

\paragraph{Generalization Evaluation Across Diverse Datasets}
We evaluate on the MS COCO-2017 dataset \cite{lin2014microsoft} to demonstrate the generalization capability of FIA-Flow (See Table~\ref{table-coco}). 
To quantify privacy leakage beyond standard IQA, we introduce the \textbf{Object Reconstruction Rate (ORR)}, which measures the consistency between the outputs of a pre-trained detector (Faster R-CNN \cite{ren2016faster}) on the original and inverted images.
Trained only on ImageNet and without fine-tuning on COCO,  FIA-Flow achieves state-of-the-art performance compared to methods that require sample-specific optimization on target features.
This cross-dataset generalization is mainly attributed to the alignment–refinement design of FIA-Flow, which learns a dataset-agnostic mapping from task features to the VAE latent space.
High ORR obtained by FIA-Flow indicates that the inverted images retain task-relevant semantics for downstream models, revealing a stronger privacy risk than IQA metrics alone capture.
The definition of ORR and the complete results are shown in the Supplementary Materials. 
\begin{table}[t]
\setlength{\tabcolsep}{1.0mm}
\renewcommand\arraystretch{0.8}
\caption{The performance comparison with different FIA methods on the COCO dataset. Bold indicates the best result of all methods.}
\centering
\label{table-coco}
\begin{tabular}{ccccc}
\toprule
 Method & LPIPS $\downarrow$ &MANIQA $\uparrow$ & $\text{ORR}_{0.5}$ $\uparrow$& $\text{ORR}_{0.75}$ $\uparrow$\\
  \midrule
 M\&V &0.700&0.5191&3.30&2.20\\
  DIP &0.332 
&0.4464&44.94&33.40\\
   SG-DIP &0.284
&0.4834&50.41&39.75\\
   FIA-Align & 0.195
&0.5981&56.02&45.84\\
  \cellcolor{myhighlight}FIA-Flow &\cellcolor{myhighlight}\textbf{0.115}&\cellcolor{myhighlight}\textbf{0.6626}&\cellcolor{myhighlight}\textbf{69.00}&\cellcolor{myhighlight}\textbf{59.33}\\
\bottomrule
\end{tabular}
\end{table}

\subsection{Ablation Studies}
\begin{figure}[t]
    \centering
    \includegraphics[width=\linewidth]{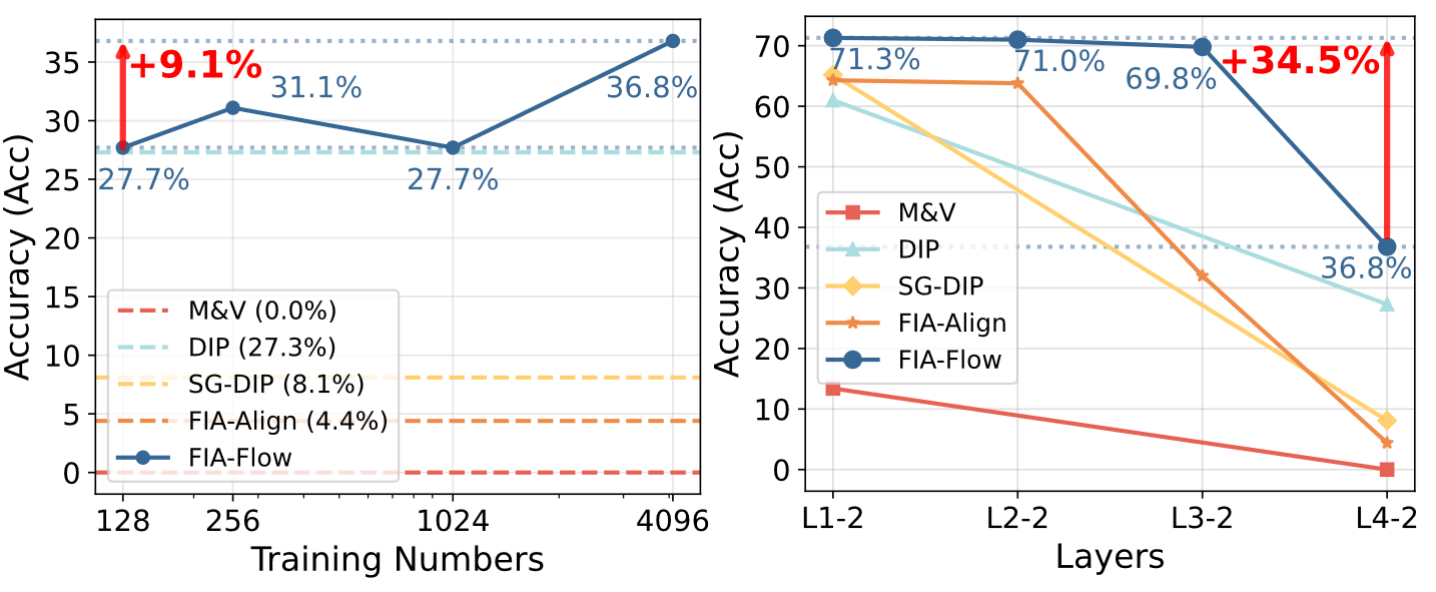}
    \caption{(a) Left: Performance comparison on the L4-2 layer with different training numbers of FIA-Flow. (b) Right: Performance comparison at different layers.}
    \label{fig-duibi}
\end{figure}

We report the main ablation results on attack-layer robustness, data efficiency, and the diffusion sampling methods and steps in the main paper. 
Additional ablations and complete results are shown in the Supplementary Materials.
\paragraph{Results on Different Training Numbers}
To test data efficiency, we trained on the ResNet-50 L4-2 layer using datasets ranging from 4,096 (0.32\%) down to just \textbf{128 (0.01\%)} samples, shown in Table \ref{table-number} and Fig.~\ref{fig-duibi}(a). 
Using only 128 samples (0.01\%), FIA-Flow not only achieves a high Acc of 27.7\% but also outperforms other methods.
The data efficiency can be attributed to LFSAM, which enforces structural alignment with the latent space through feature rearrangement that matches its dimensionality and hierarchical aggregation that reduces the mapping complexity and sample requirements.
\paragraph{Results on Different Layers}
We evaluate FIA-Flow at various depths of ResNet-50, shown in Table \ref{table-layer} and Fig. \ref{fig-duibi}(b). 
While performance degrades in deeper layers, FIA-Flow consistently outperforms other methods across all victim layers.
Notably, its performance on the deep L3-2 layer (\textbf{69.8\%} Acc) exceeds SG-DIP's 65.2\% on the shallow L1-2 layer.
Despite the loss of spatial detail in deeper layers, FIA-Flow effectively uses high-level semantic information for accurate reconstruction.
This capability underscores a serious privacy concern: FIA-Flow can recover visually detailed and semantically meaningful images from abstract representations.

\begin{table}[t]
\caption{The performance comparison of FIA-Flow with different training numbers on L4-2 of ResNet-50.}
\centering
\label{table-number}
\setlength{\tabcolsep}{1.0mm}
\begin{tabular}{ccccc}
\toprule
Number& PSNR $\uparrow$& Acc $\uparrow$& LVLM-C $\uparrow$& LVLM-PL $\uparrow$\\
\midrule
 \multirow{1}{*}{4,096(0.32\%)}  &20.31  &36.8&18.0 &0.902 \\
 \multirow{1}{*}{1024(0.08\%)} & 20.04 & 27.7&14.5 &0.898 \\
 \multirow{1}{*}{256(0.02\%)}  & 19.45 & 31.1& 12.8&0.900 \\
 \multirow{1}{*}{128(0.01\%)}  & 19.01 & 27.7&12.5 &0.898 \\
\bottomrule
\end{tabular}
\end{table}

\begin{table}[t]
\caption{The performance comparison of FIA-Flow across different victim layers of ResNet-50.}
\renewcommand\arraystretch{0.96}
\centering
\label{table-layer}
\begin{tabular}{ccccc}
\toprule
 Layer & PSNR $\uparrow$& Acc $\uparrow$& LVLM-C $\uparrow$& LVLM-PL $\uparrow$\\
  \midrule
 \multirow{1}{*}{L1-2} & 30.01 &71.3 &70.1 &0.929\\
  \multirow{1}{*}{L2-2} & 29.65 & 71.0 &69.8 &0.928\\
   \multirow{1}{*}{L3-2} & 26.29 & 69.8 &63.4 &0.913\\
  \multirow{1}{*}{L4-2} & 20.31  &36.8&18.0 &0.902 \\
 \bottomrule
\end{tabular}
\end{table}

\begin{table}[t]
\caption{The performance comparison of FIA-Flow with different sampling methods and steps on L4-2 of ResNet-50.}
\setlength{\tabcolsep}{1.0mm}
\renewcommand\arraystretch{0.7}
\centering
\label{table-step}
\begin{tabular}{cccccc}
\toprule
 Methods& Steps & PSNR$\uparrow$& Acc$\uparrow$& LVLM-C$\uparrow$& LVLM-PL$\uparrow$\\
\midrule
 \multirow{3}{*}{DDPM} & 10 &20.09 & 4.1& 5.8 &0.878 \\
  &  50 & 19.97&4.9 & 4.2& 0.876 \\
   & 200 &19.95 &4.5 & 4.8 &0.877\\
   \midrule
  \multirow{3}{*}{DIFM} &  1 &\textbf{20.31} &36.8 &18.0 &0.902 \\
  & 5 &19.61 &\textbf{38.2} &37.3 &\textbf{0.914} \\
  & 10 & 19.21&36.9 &\textbf{38.3} &\textbf{0.914} \\
\bottomrule
\end{tabular}
\end{table}

\paragraph{Results on Different Sampling Methods and Steps}
The performance gap between diffusion probabilistic model (DDPM) \cite{ho2020denoising} and DIFM reflects not only efficiency but also methodological suitability for FIA.
The iterative ``add-noise, then-denoise'' paradigm of DDPM is an indirect and stochastic process designed for diverse sampling, making it difficult for the high-fidelity reconstruction of a specific input. 
In contrast, FIA-Flow adopts a deterministic alignment-refinement paradigm, enabling efficient, high-fidelity inversion.
As shown in Table \ref{table-step}, one-step DIFM is highly effective. 
Increasing sampling steps slightly decreases PSNR but improves Acc and LVLM-based scores, suggesting increased privacy exposure.

\section{Conclusion}
In this work, we introduce FIA-Flow, a data-efficient black-box FIA framework for high-fidelity feature inversion in Split DNNs.
Benefiting from the alignment-refinement strategy, FIA-Flow significantly outperforms state-of-the-art methods, especially in recovering details across diverse architectures and layers.
FIA-Flow's effectiveness and data efficiency demonstrate that Split DNNs face a more severe and practical privacy threat than previously recognized.
These findings underscore the urgent need for designing robust and efficient defense mechanisms that can mitigate privacy risks while preserving model utility and inference performance.
\par

{
    \small
    \bibliographystyle{ieeenat_fullname}
    \bibliography{main}
}
\end{document}